\documentclass{bmvc2k}

\title{Interactive Video Retrieval with Dialog}

\addauthor{Sho Maeoki}{maeoki@mi.t.u-tokyo.ac.jp}{1}
\addauthor{Kohei Uehara}{uehara@mi.t.u-tokyo.ac.jp}{1}
\addauthor{Tatsuya Harada}{harada@mi.t.u-tokyo.ac.jp}{12}

\addinstitution{
Graduate School of Information Science and Technology, \\
The University of Tokyo \\
Tokyo, Japan
}
\addinstitution{
RIKEN
}

\runninghead{MAEOKI, UEHARA, HARADA}{INTERACTIVE VIDEO RETRIEVAL WITH DIALOG}

\def\etal{\emph{et al}\bmvaOneDot}

\newcommand{\figref}[1]{Fig.~\ref{#1}}
\newcommand{\tabref}[1]{Table~\ref{#1}}
\newcommand{\equref}[1]{Eq.~\ref{#1}}
\begin{document}

\maketitle

\begin{abstract}
Now that everyone can easily record videos, the quantity of which is continuously increasing, research on methods for improved video retrieval is important in the contemporary world. In cases where target videos are to be identified within a large collection gathered by individuals, the appropriate information must be obtained to retrieve the correct video within a large number of similar items in the target database. 
The purpose of this research is to retrieve target videos in such cases by introducing an interaction, or a dialog, between the system and the user. 
We propose a system to retrieve videos by asking questions about the content of the videos and leveraging the user's responses to the questions.
Additionally, we confirmed the usefulness of the proposed system through experiments using the dataset called AVSD which includes videos and dialogs about the videos.

\end{abstract}

%------------------------------------------------------------------------- 
\section{Introduction}
\label{sec:intro}
Nowadays, with the widespread use of smartphones anyone can easily record videos, leading to an ever-increasing amount of content.
It can be said that research on video retrieval has great significance.
Videos taken by individuals, including home videos and life log videos, generally do not become highly popular and are not distinguished in most cases. Therefore, we cannot use elements such as the number of hits and tag data, which can be used when retrieving videos on the web; this makes retrieval difficulty relatively high.  However, there is a motivation for retrieving such videos. For example, there are situations where videos taken in the past are very impressive and valuable, but they are buried in other videos taken in the past and cannot be found easily.
The premise of this research is to retrieve target videos in this type of situation.

\begin{figure}[htbp]
 \centering
 %\fbox{\rule{0pt}{2in} \rule{.9\linewidth}{0pt}}
 %\includegraphics[width=0.8\textwidth,height=60mm]{images/intro.pdf}
 \includegraphics[width=0.8\textwidth]{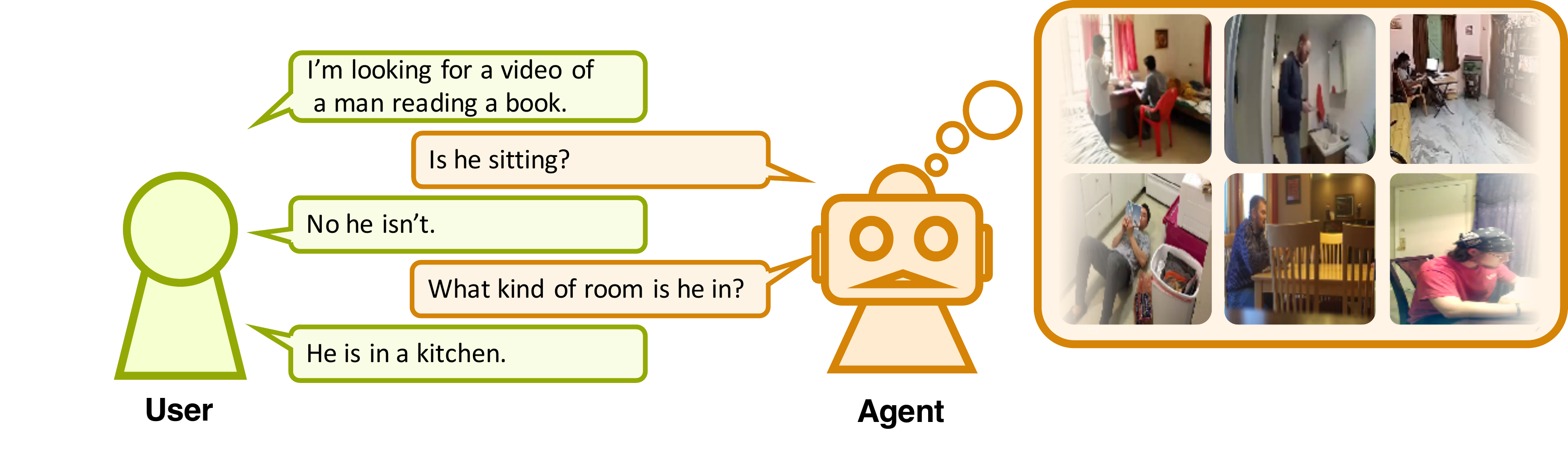}
 \caption{Outline of video retrieval with dialog. The picture shows some frames of videos of "a man reading a book". It is assumed that there are various situations for the same behavior. It is possible to narrow down the candidates making use of appropriate questions and their responses.}
 \label{fig:reading}
\end{figure}

To facilitate the retrieval of target videos among similar items, this study proposes a retrieval technique using interaction as shown in \figref{fig:reading}. For example, when a retrieval is requested using the sentence "a man reading a book" as shown in \figref{fig:reading}, many similar videos are valid candidates. It is necessary to add more information to distinguish the videos. However, in reality, it is difficult to assume that a user is aware of the sentence that is suitable enough to represent the videos they are searching for, or it is labor-intensive. Indeed, under the current assumption, the user has recorded a large number of similar videos, and we can consider that it is impossible to remember the details of all videos. Therefore, we propose introducing a dialog as shown in \figref{fig:reading}. The agent in the system asks a question to efficiently search for videos that the user wants, and then the user replies to the questions. 
%We call the part of the system that controls the interaction with the user the "agent".
For example, suppose that there are six videos in \figref{fig:reading} and we try to retrieve a video of "a man reading a book" in \figref{fig:reading}. If the search is for the upper left video, the agent should ask "How many people are in the video?". And if asking "Is the person reading a book while lying down?", the lower left video can be separated from the others in \figref{fig:reading}. If you ask the question "Does the person read while standing", you can search for the video in the upper middle of \figref{fig:reading}. You can also distinguish videos by asking questions such as "What was the person doing at the start of the video?" and "What was the person doing before (after) reading a book?". Moreover, since the user is assumed to have some knowledge of the video he is searching for, there is no need to watch videos when answering these questions. That is to say, ideally the user does not have to look at the displayed candidate videos and identify the optimal query and can instead simply search for the target video by answering the questions from the agent.
It is manifest that video retrieval can be performed effectively by introducing this type of dialog.

The purpose of this research is to retrieve target videos among similar items by introducing a dialog between the system and the user.
The contributions of this research are as follows. (1) We proposed a new video retrieval task that utilizes interactive elements (i.e., dialog) and implemented a model to instantiate it. (2) We clarified the role of each module in this method; the method for embedding features has an effect on retrieval performance itself, and the way of encoding dialog history influences the relationship between the retrieval performance and the progress of dialog. (3) We conducted the user study and confirmed that the method was effective for the task proposed in this study.

%------------------------------------------------------------------------- 

%------------------------------------------------------------------------- 

%------------------------------------------------------------------------- 

%------------------------------------------------------------------------

\section{Related Work}
\subsection{Text-based Video Retrieval}\label{sec:rw_retrieval}
In the video retrieval method using text, we first learn a mapping that transforms text and video features into a joint embedding space \cite{wang2016comprehensive}. Then, in the learned joint embedding space, a video with a high degree of similarity to the sentence used as the input query is output as a search result. In the past, canonical correlation analysis (CCA) has been used as an approach to learning the mapping of the joint space. Learning was done to maximize the covariance of the distribution of the two different modalities in the embedding space.
Presently, methods using deep neural networks (DNN) are popular \cite{yu2018joint,mithun2018learning,miech18learning,yamaguchi2017spatio,shao2018find,yang2018text2video,otani2016learning,dong2016word2visualvec,Hendricks2017LocalizingMI}.
There are two types of features embedded in the joint space in the case of video retrieval, a sentence feature and a video feature. The former often inputs text such as captions to recurrent neural networks (RNN) that can handle time-series data and adopts its final hidden state as the representation. The latter is effective when considering features obtained by applying convolution neural networks (CNN) to each frame of a video in a multilateral manner \cite{yu2018joint, mithun2018learning, miech18learning, yamaguchi2017spatio} .
 In these related studies, basically, the corresponding video is output with a short sentence of approximately one sentence as input, and the structure for handling the dialog history which includes a plurality of sentences cannot be deemed sufficient. For this reason, this cannot be used in this research, which has to deal with dialog.
 
\subsection{Vision and Dialog}
The visual dialog proposed by Das \etal{}~\cite{visdial} is a task that takes an image and multiple questions as inputs and subsequently outputs a response to each question. Based on this, visual dialog is still actively researched, and research on video dialog that targets videos instead of images has begun \cite{zhao2018multi,hori2018end,jang-CVPR-2017,pasunuru2018game,alamri2019audiovisual}.
However, these studies are aimed at returning better responses based on the contents of the videos, and there is no module for video retrieval.
 
In contrast, there are also studies that have proposed training methods for the goal-oriented visual dialog. Das \etal{}~\cite{das2017visdialrl} enables interactive image retrieval with dialog. They generate an asymmetric scene in which an image can be viewed from an answerer, while a questioner cannot see the image. Under the circumstances, the questioner asks a question about the image to the answerer. The answerer in turn gives a response so that the questioner can gain a finer understanding of the corresponding image.
%At the time of training, the questioner tries to regress the image using the dialog history without knowing the image feature of the GT known to the answerer. They argue that it is possible to improve dialog performance by performing collaborative reinforcement learning on this task.
As a method for evaluating the questioner, Das \etal{}~\cite{das2017visdialrl} proposed an image retrieval method using image features obtained by regression from the dialog history representation. This can be interpreted as an image retrieval with dialog.
However, Das \etal{}~\cite{das2017visdialrl} only considered the point that the image feature predicted by the questioner approaches the feature of the GT image known to the answerer in training. Therefore, it is impossible to distinguish between the target video and similar ones. Thus, we consider that this is not sufficient to achieve the purpose of this research.

\section{Model}
%\subsection{Functional Requirement}
The following three function requirements (FR) are considered necessary to achieve the goal of this research.
%Three functional requirements (FR) are considered necessary to achieve the goal of this research.

%\noindent{\bf FR1: Making use of user's responses and dialog history.}
\noindent{\bf FR1: Making use of dialog history.}
First, to utilize of the dialog, the proposed system needs to be able to effectively use of user responses and of the history of dialog. For example, a response to a question is provided, the dialog history must be used adequately to improve retrieval performance. In addition, when generating an effective question to identify a target video, it is necessary to generate a question on information that is not known yet.

\noindent{\bf FR2: Taking spatio-temporal elements of videos into account. }
Second, in this research, the search targets are not static images but videos. Therefore, we need to handle concepts that cannot be understood solely by looking at the images (only the image features of the 2-D CNN) but instead by seeing the videos in the flow. 
%For example, if someone has an open book on their desk and they have an image with a man holding it with both hands, we usually think that he is reading a book, but it may be a frame before he leaves somewhere to put the book on the desk.
For example, there are some cases we cannot tell whether a person in a video is walking or just standing by looking at a frame of the video. It can happen as the frame lacks dynamic elements of the video.
This is a spatio-temporal element that cannot be handled without using features for action recognition, such as 3-D CNN.

\noindent{\bf FR3: Generating appropriate questions based on videos.}
Third, to narrow down the target videos among similar items by dialog, the system needs to generate appropriate questions. If a question that is completely off the mark is generated, the retrieval performance will not improve, and the dialog will be meaningless. Therefore, the questions generated need to adapt based on the videos searched.
\subsection{Modeling and Overview}

\begin{figure*}
\begin{center}
%\fbox{\rule{0pt}{2in} \rule{.9\linewidth}{0pt}}
%\bmvaHangBox{\fbox{\includegraphics[width=5.6cm]{images/model_overview.pdf}}}
\includegraphics[width=0.7\textwidth]{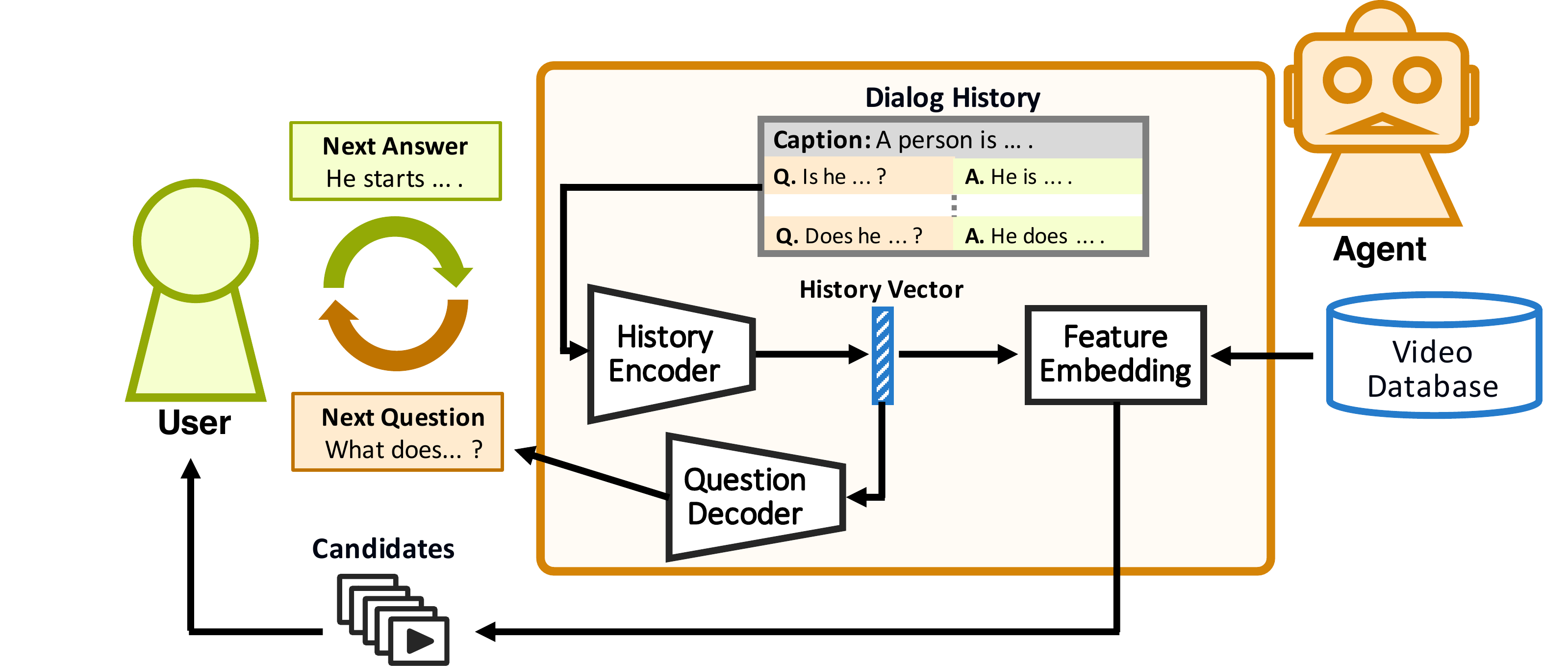}
\end{center}
   \caption{An overview of the proposed model and each module.}
\label{fig:behavior}
\end{figure*}

In this section, we provide an overview of a model of the proposed method that satisfies the functional requirements mentioned above. \figref{fig:behavior} is an overview.
The following describes the proposed system, assuming that one combination of a question and its response is called one round of dialog. As shown in \figref{fig:behavior}, in the proposed system there is an agent interacting with the user. The agent's main role is to generate the question and present candidate videos based on the dialog while interacting with the user. In this system, the user first inputs a query describing the video that he is searching for, and the agent outputs a question in return.
The proposed system starts from the point where the user inputs a natural sentence describing the target video to the agent. When the first sentence $D_{0}$ is input, the agent uses that sentence as a query, and presents to the user several ($N$) videos that are close to the query in the feature space, namely ${\rm Cand}_{0}$ = $\{ C^{(1)}_{0}, C^{(2)}_{0}, ..., C^{(N)} _{0} \}$. In this study, we assume $N = 10$.  It is defined as the 0th round of the dialog until this first sentence $D_{0}$ is input and the first candidate videos ${\rm Cand}_{0}$ are output. After completing the 0th round dialog, the agent generates a question $q_{1}$, and the next round of the dialog begins based on this. Considering the t-th round dialog ($t = 1, ...,  T$), the question $q_{t}$ generated after the $(t-1)$-th round dialog and the user's response $a_{t}$ consist of the $t$-th round dialog. That is, if the pair of the question and the response composing the $t$-th round dialog is written as $H_{t} = [q_{t}, a_{t}]$, the series of processes are expressed as follows.
 \begin{eqnarray}
  {\bf s}_{t} &=& {\rm HistEnc} (D_{0},H_{1},...,H_{t}\,;\,\theta_{HE}) \label{eq:make_hist} \\
  {\rm Cand}_{t} &=& \mathop{{\rm top}\,N }_{{\bf v} \in V}\, S(\,{\rm DialogEmb}({\bf s}_{t}\,;\,\theta_{DE}),{\rm VideoEmb}(V\,;\,\theta_{VE}))
  \label{eq:make_cand}\\
  q_{t+1} &=& {\rm QuesDec}(\,{\bf s}_{t}\,;\,\theta_{QD})
  \label{eq:gen_ques} 
 \end{eqnarray}

 HistEnc$(D_{0}, H_{1}, ..., H_{t}) $ in \equref{eq:make_hist} is a mapping to obtain the representation of the dialog history ${\bf s}_{t}$ until the $t$-th round. In the following, we call the representation of the dialog history $ {\textbf s}_{t}$ the "history vector". \equref{eq:make_cand} represents the mapping that uses the history vector ${\textbf s}_{t}$ obtained by \equref{eq:make_hist} to obtain $N$ candidate videos $ {\rm Cand}_{t} $ from the video feature groups $ V $ in the database.  This mapping is composed of DialogEmb ($ \cdot $) and VideoEmb ($ \cdot $) which embed the history vector ${\textbf s}_{t}$ and the feature group $ V $ respectively into the joint space. In the joint embedding space, the similarity function $ S (\cdot, \cdot) $ calculates the closeness between the embedded history vector and the embedded video features, and as a result $N$ items considered to have a high similarity to the embedded history vector are selected. QuesDec $(\cdot)$ in \equref{eq:gen_ques} is a mapping that outputs the next question $q_{t + 1}$ with the history vector $ {\textbf s}_{t} $ as an input. 
\equref{eq:make_hist}, \equref{eq:make_cand}, and \equref{eq:gen_ques} are the history encoding module (History Encoder), the feature embedding module (Feature Embedding) and the question generation module (Question Decoder) in \figref{fig:behavior}. All modules are executed based on $\theta_{HE},\theta_{DE},\theta_{VE},\theta_{QD}$.
In this study, the first sentence $D_{0}$ is considered as a caption of a video, and the end is assumed to be the point where the round of dialog reaches a predetermined upper limit $ T (= 10) $.

\subsection{Model Learning}
{\bf History Encoder.}
This module is responsible for encoding dialog history. The architecture of this module is similar to the Hierarchical Recurrent Encoder in  Das \etal{}~\cite{das2017visdialrl}.
The dialog in each round is decomposed into words and then represented as a vector by the word embedding matrix to obtain sentence features in order. Then it is input in an LSTM (Sentence Encoder).
Let $ {\textbf F}_{0}, ..., {\textbf F}_{t} $ be the output obtained by the Sentence Encoder.
These sentence-level features are input sequentially in an LSTM (State Encoder) which is different from the Sentence Encoder. The final hidden state obtained from the State Encoder is considered to be a feature that represents the entire dialog history effectively. In this study, for the history vector ${\bf s}_{t}$, we adopted the concatenation of the final hidden state of the State Encoder and the feature of the caption which is the first output of the State Encoder. This is because in the task proposed in this study, the caption of the video holds the largest amount of information and the subsequent dialog is considered to supplement the contents of the caption.
 The history vector $ {\textbf s}_{t} $ obtained as described above is a vector that semantically reflects the history of past conversations. Therefore, using this feature as an input to construct a later pipeline will satisfy FR1.

 \noindent{\bf Feature Embedding.}
 In this module, videos are searched using the dialog history as a query, and candidate videos are output.
The history vector ${\bf s}_{t}$ obtained by \equref{eq:make_hist} is used as an input of this module, which is responsible for embedding into the joint space.
%Embedding the history vector ${\bf s}_{t}$ into the joint space is interpreted as an analogy to embedding text features and video features into the joint space in video retrieval using natural language queries.
To reduce the computational cost at the time of training, the video feature in this study uses the one that has been previously extracted using a pretrained model and pooled in the time direction. Let the extracted video features be $ {\bf v}$, and the history vector be $ {\textbf s}_{t}$.
In the joint embedding space, training proceeds by minimizing the following loss function \equref{eq:ranking_loss} according to the research of Mithun \etal{}~\cite{mithun2018learning} .
 \begin{equation}
  \label{eq:ranking_loss}
  \sum_{v}L(r_{v})\, {\rm max}(0,\alpha - S(\overline{\bf v},\overline{{\bf s}}_{t}) +S(\overline{\bf v},\hat{{\bf s}}_{t})) + \sum_{s_{t}}L(r_{s})\, {\rm max}(0,\alpha - S(\overline{{\bf s}}_{t},\overline{\bf v}) + S(\overline{{\bf s}}_{t},\hat{\bf v}))
 \end{equation}
 $\alpha$ in \equref{eq:ranking_loss} is a margin and $S(\cdot,\cdot)$ is a function for similarity calculation. In this research, cosine similarity is adopted as $S(\cdot,\cdot)$, as in the study of Mithun \etal{}~\cite{mithun2018learning}.
 Note that $L(r) = 1 + 1/(N_{v}-r+1)$, where $N_{v}$ is the number of compared videos. $r_{v}$ and $r_{s}$ in $L(r)$ indicate the ranks at which GT videos and dialogs are located in the batch. For a video embedding $\overline{{\bf v}} = {\rm VideoEmb(v)}$, $r_{v}$ is the rank of the matching dialog ${\bf s}_{t}$ among all comparisons. Similarly, for an embedding of the history vector $\overline{{\bf s}}_{t}={\rm DialogEmb}(\textbf{s}_{t})$, $r_{s}$ is the rank of the matching video among all comparisons. 
 Linear layers are adopted for VideoEmb$(\cdot)$ and DialogEmb$(\cdot)$.
 %as ${\bf \overline{v}} = W_{v}{\bf v}+{\bf b_{v}},\; \overline{{\bf s}_{t}} = W_{s}{\bf s}_{t}+{\bf b_{s}}$.
 %Note $W_{v}\in {\rm R}^{D \times V},\,
 %  W_{s}\in {\rm R}^{D \times S},\,
 %  {\bf b_{v}} \in {\rm R}^{D},\,
 %  {\bf b_{s}} \in {\rm R}^{D}$.
 %  $W_{v},W_{s}$ and ${\bf b_{v}},{\bf b_{s}}$ are weight parameters and bias parameters optimized by training.
 $\hat{\bf v}$ and $\hat{{\bf s}}_{t}$ are called hard negatives (i.e., the negative video/dialog sample closest to a positive matching $(\overline{{\bf s}}_{t},\overline{\bf v})$ pair).
 Accordingly training is supposed to be done to learn to maximize the similarity between video embeddings and the corresponding dialog embeddings, and minimize similarity to hard negatives.
 In this module, the history vector representing the dialog history and the video features are directly related in the joint space. If the video features reflect the spatio-temporal elements of the videos, the dialog history and the spatio-temporal elements of the videos are considered linked, consequently FR2 is satisfied.

 \noindent{\bf Question Decoder.}\label{sec:ques_dec}
 This module is responsible for the question generation necessary to retrieve videos with dialog.
 The history vector ${\bf s}_{t}$ is input as the initial hidden state of an LSTM that composes the module of \equref{eq:gen_ques}. The hidden state of the LSTM is used to calculate the probability of the words in the generated question.
 While training, we perform supervised learning with teacher forcing using cross entropy loss. Therefore, assuming that the dataset has questions of a nature that satisfies the functional requirement, we can expect to be able to generate appropriate questions. In other words, this module supports FR3.

 \noindent{\bf Summary.}
 When training the whole model, we minimize these linear sums by writing the Question Decoder's cross entropy loss as DialogLoss, and the loss function in the Feature Embedding shown in \equref{eq:ranking_loss} as FeatLoss. The coefficients $ a, b $ in the equation \equref{eq:total_loss} are hyperparameters that indicate the extent to which each module is emphasized.
 \begin{equation}
 \min_{\theta}\, \left(a\,{\rm DialogLoss} + b\,{\rm FeatLoss}\right) \label{eq:total_loss}
 \end{equation}

\section{Experiments}
\subsection{Settings}
{\bf Dataset.}
In this study, we used the AVSD dataset \cite{alamri2019audiovisual}.
This dataset was created by adding dialog data to the existing video dataset called Charades \cite{sigurdsson2016hollywood}. Charades is a video dataset of about 30 seconds, and the collection continues with activities that people will be exposed to in their daily lives. Charades has many motions in one video and is characterized by the presence of many semantically similar videos in the dataset.
The AVSD dataset contains ten rounds of questions and answers for each video. Questions include a lot of spatio-temporal information (e.g., the development of events).
From the above, it can be said that the AVSD dataset, which includes dialogs focusing on elements unique to the videos, targeting recordings by individuals including home videos and lifelog videos, is suitable for this research.
%\tabref{tab:sample_num} shows the number of samples of the dataset actually used. 
AVSD has 7,985 samples for training. We used 863 samples for validation and 1,000 samples for testing.
Part of the validation data was adopted as test data in this study.

%\begin{table}[htb]
%  \caption{The number of the dataset samples.}
%  \centering
%  \begin{tabular}{lccc}\hline
%    & training data & validation data& test data\\ \hline 
%    the number of videos& 7985  & 863 & 1000  \\ \hline
%  \end{tabular}
%  \label{tab:sample_num}
%\end{table}

\noindent{\bf Implementation Details.}\label{sec:setting}
We prepare two types of features for the representation of videos, the feature extracted for each frame with ResNet152 \cite{he2016deep} pretrained with ImageNet \cite{deng2009imagenet} and the feature extracted from the I3D \cite{carreira2017quo} model pretrained with Kinetics \cite{kay2017kinetics}. Furthermore, the VGGish \cite{Hershey2017CNNAF} feature pretrained on the Audio Set \cite{gemmeke2017audio} is prepared as an audio feature for supplementary data. All these features are pooled in the time direction to use.
Moreover, since the model proposed in this research is a complex configuration with multiple modules, it is difficult to train all parameters end-to-end at once. Therefore, training is divided into two steps; in the first step, the parameters of VideoEmb ($ \cdot $) and DialogEmb ($ \cdot $) are fixed to the initial values, and in the second step, this constraint is released to train all parameters. The main hyperparameters are set as follows. The batch size is 32, the word embeddings are 300-d, the hidden state of the two LSTMs in the History Encoder is 512-d, the hidden state of an LSTM in the Question Decoder is 1,024-d, the dimension of the joint embedding space is 1,024, $\alpha$ in \equref{eq:ranking_loss} is 0.2 and the coefficients in \equref{eq:total_loss} are $a = 2, b = 1,000$. Parameters are optimized using Adam \cite{kingma:adam} with an initial value of 0.001. The number of dimensions of the video feature are ResNet152: 2,048, I3D: 1,024, VGGish: 128.

\noindent{\bf Evaluation Metrics.}
Here we introduce the evaluation metrics used in this research. We measure the rank-based performance by Recall@$k$ (R@$k$) and Mean Rank (MeanR). R@$k$ calculates the percentage of the test samples for which the GT video is found in the top-$k$ retrieved points and MeanR calculates the mean rank of all GT videos. Higher is better for R@$k$ and lower is better for MeanR.
%We report R@1, R@5, and R@10 in addition to MeanR.
Note that R@$k$ ($k=1, 5, 10$) and MeanR in \tabref{tab:feat_table} and \tabref{tab:comparison} are the values obtained when 10 rounds of GT dialog data are input.

\noindent{\bf Baselines.}
 As baselines, we prepare three types of models overall, namely the basic model, L2 Loss, and LSTM. The basic model adopts the final hidden state of the State Encoder in the History Encoder as a history vector ${\textbf s}_{t}$ without performing a concatenation of the caption feature. The other two baselines are based on this basic model. 
 In the L2 Loss model, L2 Loss is applied in Feature Embedding instead of ranking loss in \equref{eq:ranking_loss}. L2 Loss centers on bringing the positive samples closer with no consideration for keeping the negative samples apart.
 The LSTM model is a model which uses an LSTM as History Encoder. In the basic model each round of dialog is encoded as a sentence feature by LSTM (Sentence Encoder), and the history vector ${\textbf s}_{t}$ is extracted hierarchically by inputting these sentence features into the LSTM (State Encoder) again. Meanwhile in the LSTM model the dialog history is input in the same LSTM.

\subsection{Results and Discussions}
{\bf Feature Selection for Representing Videos.}
 Here we select the appropriate representation for videos. We compared the retrieval performance with the ResNet feature, I3D feature and VGGish feature. After that, experiments were also conducted with various types of features combined hoping that multiple features could be used effectively. This is shown in \tabref{tab:feat_table}. We adopt max pooling for pooling features.
 %In addition, since max pooling showed higher performance than average pooling for any of the three features, max pooling is adopted as the pooling method here.
According to \tabref{tab:feat_table}, it can be said that the best performance is achieved by adopting a combination of the I3D feature and VGGish aside for the MeanR. In the following, this combination is adopted as the video representation.
Comparing I3D and ResNet, I3D gives better results because the dataset used this time has a more dynamic behavior, and I3D is more likely to reflect such features. The audio feature (VGGish) is poor in terms of stand-alone performance, but it contributes to the improvement of the retrieval performance when considered simultaneously with I3D. However, as per the lower three columns in \tabref{tab:feat_table}, it is apparent that simply combining them does not necessarily lead to an improvement in retrieval performance.
\begin{table}[htb]
  \centering
  \caption{Comparison of retrieval performance depending on the input features. "+" in the table means concatenation.}
  \begin{tabular}{lcccc}\hline
     & R@1 & R@5 & R@10 & MeanR\\ \hline \hline
    %I3D(Ave) & 3.00  & 10.8 & 17.3 & 147\\ \hline
    I3D & {\bf 4.20} & 13.1 & 21.6 & {\bf 116}\\ \hline
    %ResNet(Ave) & 1.00 & 4.70 & 9.20 & 220\\ \hline
    ResNet & 1.70 & 6.1 & 10.0 & 190\\ \hline
    %VGGish(Ave) & 0.500 & 1.80 & 3.00 & 408\\ \hline
    VGGish & 0.00 & 1.60 & 3.90 & 383\\ \hline
    I3D + ResNet + VGGish & 3.60 & 11.9 & 18.6 & 147\\ \hline
    ResNet + VGGish & 1.60 & 6.80 & 11.0 & 204\\ \hline
    I3D + VGGish & {\bf 4.20} & {\bf 13.5} & {\bf 22.1} & 119\\ \hline
  \end{tabular}
  \label{tab:feat_table}
  
\end{table}

\begin{table}[htb]
  \caption{Comparison of retrieval performance against baselines. The basic model and the proposed model achieve much better performance than the L2 Loss and the LSTM. The proposed model marks performance improvement in R@$k$ from the basic model. }
  \centering
  \begin{tabular}{lcccc}\hline
     & R@1 & R@5 & R@10 & MeanR\\ \hline \hline
    L2 Loss & 0.600 & 2.40 & 4.20 & 377 \\ \hline
    LSTM & 0.500 & 2.30 & 5.40 & 371\\ \hline
    Basic Model  & 2.90 & 9.40 & 16.8 & 123\\ \hline
    Proposed Model  & {\bf 4.20} & {\bf 13.5} & {\bf 22.1} & {\bf 119}\\ \hline
  \end{tabular}
  \label{tab:comparison}
\end{table}

%\begin{figure}
%\begin{tabular}{cc}
%{\includegraphics[width=4.5cm]{images/meanr.png}}&
%{\includegraphics[width=4.5cm]{images/r10.png}}\\
%(a)&(b)
%\end{tabular}
%\caption{Transition of MeanR (a) and R@10 (b) as the dialog proceeds. GT dialog data is used as input. The error bar in MeanR means the standard error. }
%\label{fig:meanr}
%\end{figure}

\begin{figure}[htbp]
  \begin{center}
    \begin{tabular}{c}

      % 1
      \begin{minipage}{0.4\hsize}
        \begin{center}
          \includegraphics[clip, %width=4.5cm]{images/meanr.png}
          width=5.0cm]{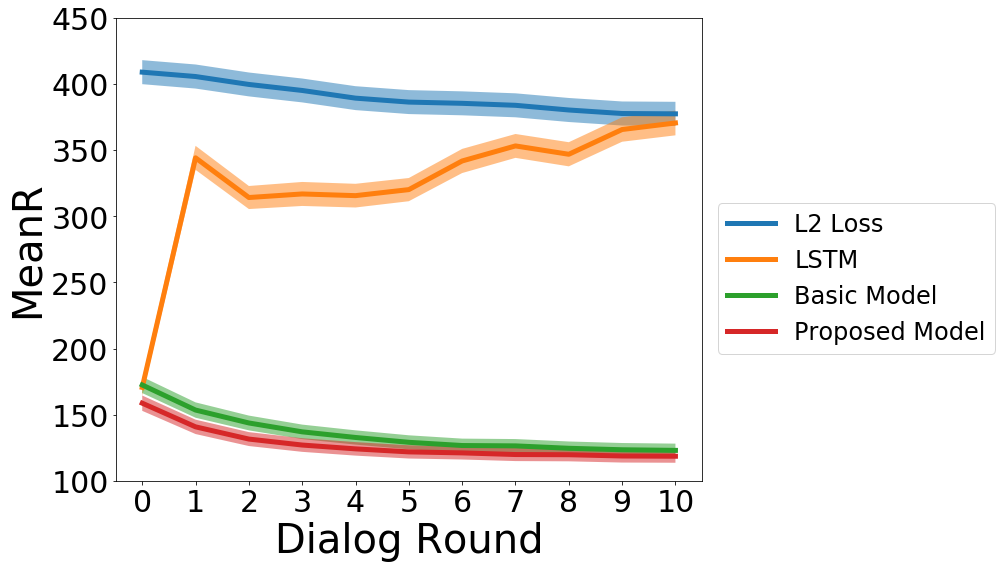}
          \hspace{1.6cm} (a)
        \end{center}
      \end{minipage}

      % 2
      \begin{minipage}{0.4\hsize}
        \begin{center}
          \includegraphics[clip, %width=4.5cm]{images/r10.png}
          width=5.0cm]{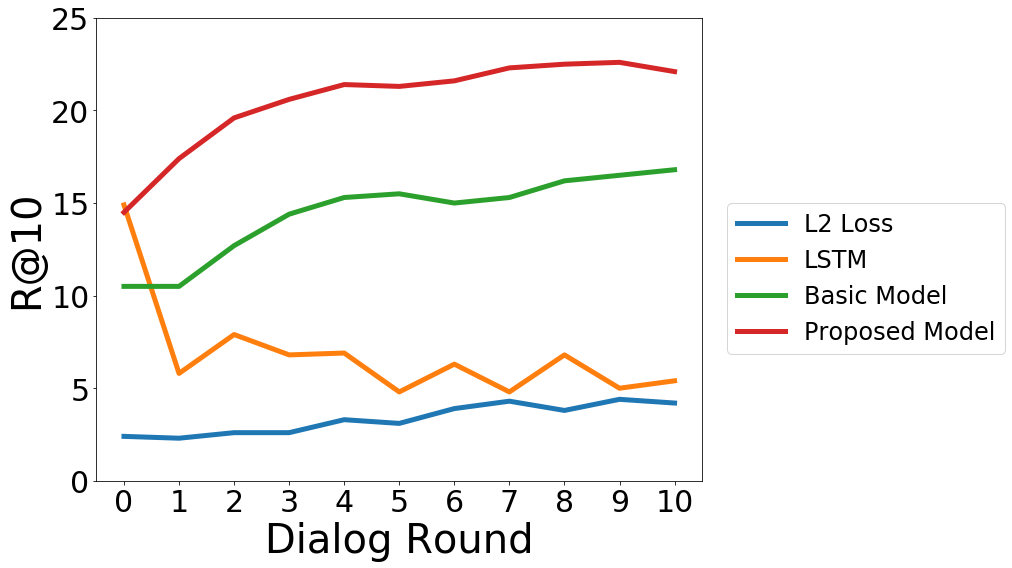}
          \hspace{1.6cm} (b)
        \end{center}
      \end{minipage}

    \end{tabular}
    \caption{Transition of MeanR (a) and R@10 (b) as the dialog proceeds. GT dialog data is used as input. The error bar in MeanR means the standard error.}
    \label{fig:meanr}
  \end{center}
\end{figure}

\noindent{\bf Comparisons against Baselines.}
\tabref{tab:comparison} shows the performance comparison against baselines. As can be seen in \tabref{tab:comparison}, we can confirm that the proposed method is superior for all metrics.
Furthermore, \figref{fig:meanr} expresses the relationship between the number of dialog rounds and the retrieval performance. It is apparent that in the proposed method, MeanR tends to decrease as the dialog progresses. As for R@10, the progress of the dialog and the performance improvement are linked, and the significance of the dialog is apparent.
First, we will compare the basic model with two methods, namely L2 Loss and LSTM. According to \figref{fig:meanr}, it appears that the L2 Loss tends to improve retrieval performance as the dialog progresses, as does the basic model. However, the retrieval performance itself is poor. This indicates that the loss function in Feature Embedding is insufficient with an L2 Loss. In other words, we can conclude that the retrieval performance itself can be improved by devising embedding methods in the joint space in Feature Embedding. Meanwhile, looking at \figref{fig:meanr}, although the LSTM achieves almost the same performance as the basic model in the 0th round, its performance deteriorates drastically in the first round and generally deteriorates thereafter. We consider that this phenomenon occurs because the LSTM cannot effectively handle a long-term information sequence such as a dialog history. In other words, it is possible to guarantee an improvement in the search performance along with the progress of the dialog by finding a way to handle a long-term information sequence in the History Encoder.
Finally, we compare the proposed model and the basic model. According to \figref{fig:meanr}, there is no big difference in MeanR between the two models, but for R@10, the proposed model is consistently superior. As a result, we can confirm that the priority of the target videos can be raised by emphasizing the caption in this task.

%\begin{figure}
%\begin{tabular}{cc}
%{\includegraphics[width=4.5cm]{images/user_study_meanr.png}}&
%{\includegraphics[width=4.5cm]{images/user_study_r10.png}}\\
%(a)&(b)
%\end{tabular}
%\caption{Transition of MeanR (a) and R@10 (b) as dialog proceeds when user study is performed with the proposed method. The error bar in MeanR means the standard error.}
%\label{fig:user_study}
%\end{figure}

\begin{figure}[htbp]
  \begin{center}
    \begin{tabular}{c}

      % 1
      \begin{minipage}{0.4\hsize}
        \begin{center}
          \includegraphics[clip, %width=4.5cm]{images/user_study_meanr.png}
          width=3.8cm]{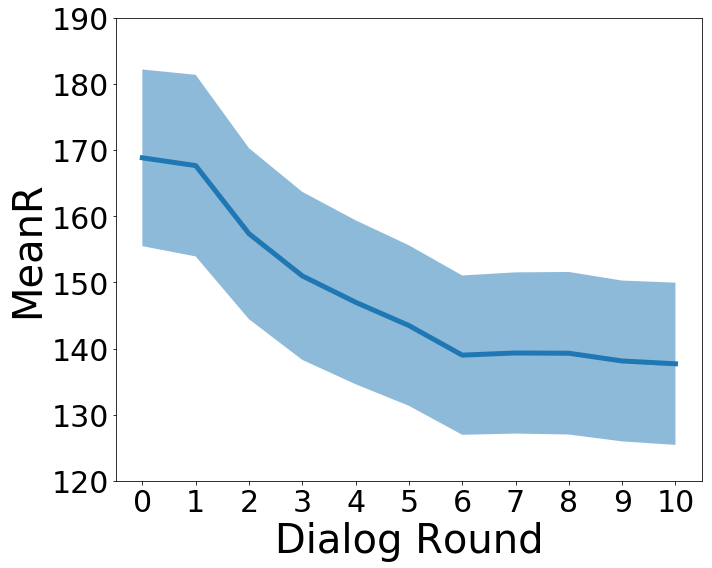}
          \hspace{1.6cm} (a)
        \end{center}
      \end{minipage}

      % 2
      \begin{minipage}{0.4\hsize}
        \begin{center}
          \includegraphics[clip, %width=4.5cm]{images/user_study_r10.png}
          width=3.8cm]{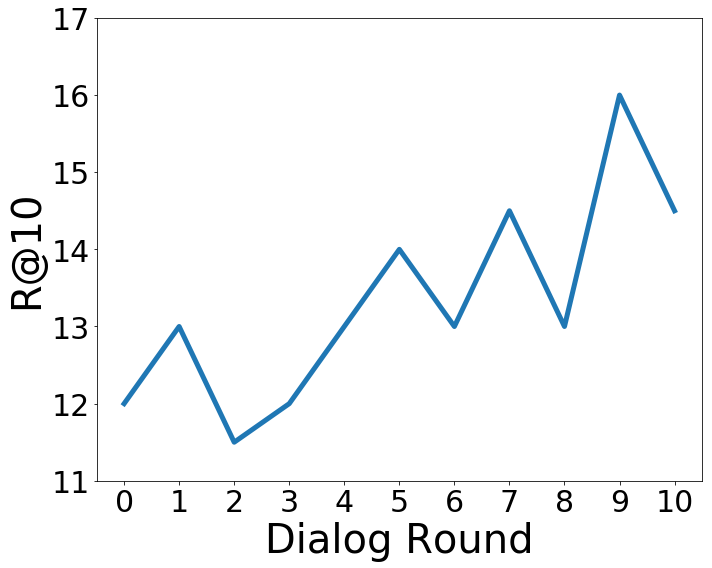}
          \hspace{1.6cm} (b)
        \end{center}
      \end{minipage}

    \end{tabular}
    \caption{Transition of MeanR (a) and R@10 (b) as dialog proceeds when user study is performed with the proposed method. The error bar in MeanR means the standard error.}
    \label{fig:user_study}
  \end{center}
\end{figure}

\noindent{\bf User Study Results.}
In the mechanical evaluation, the dialog with humans is simulated using GT dialog data as input. However, an evaluation through a human-model interaction is also necessary.
Therefore, we conducted a user study using Amazon Mechanical Turk (AMT) to confirm that the retrieval performance improves with the dialog even when actually interacting with humans.
We randomly selected 200 out of 1,000 videos of test data.
Considering the purpose of this research, as a way of user study, ideally users should provide their own videos, and hold the conversations searching for the videos. However, such an evaluation method is difficult in practice. For this reason, we used the videos in the dataset as the targets and performed video retrieval with dialog. 
As a specific procedure, we asked workers to respond to questions for ten rounds based on the content of the videos and their captions. The ranks of the GT videos in the database were recorded using the dialog data obtained.
Looking at \figref{fig:user_study}, it appears that the retrieval performance improves with the progress of the dialog for both MeanR and R@10, as in the case involving the GT dialog data.
Therefore, we confirmed that the improvement of the retrieval performance itself can be enhanced using dialog even when a dialog with a human is actually performed.

\begin{figure}[htbp]
 \centering
  \includegraphics[width=0.8\textwidth]{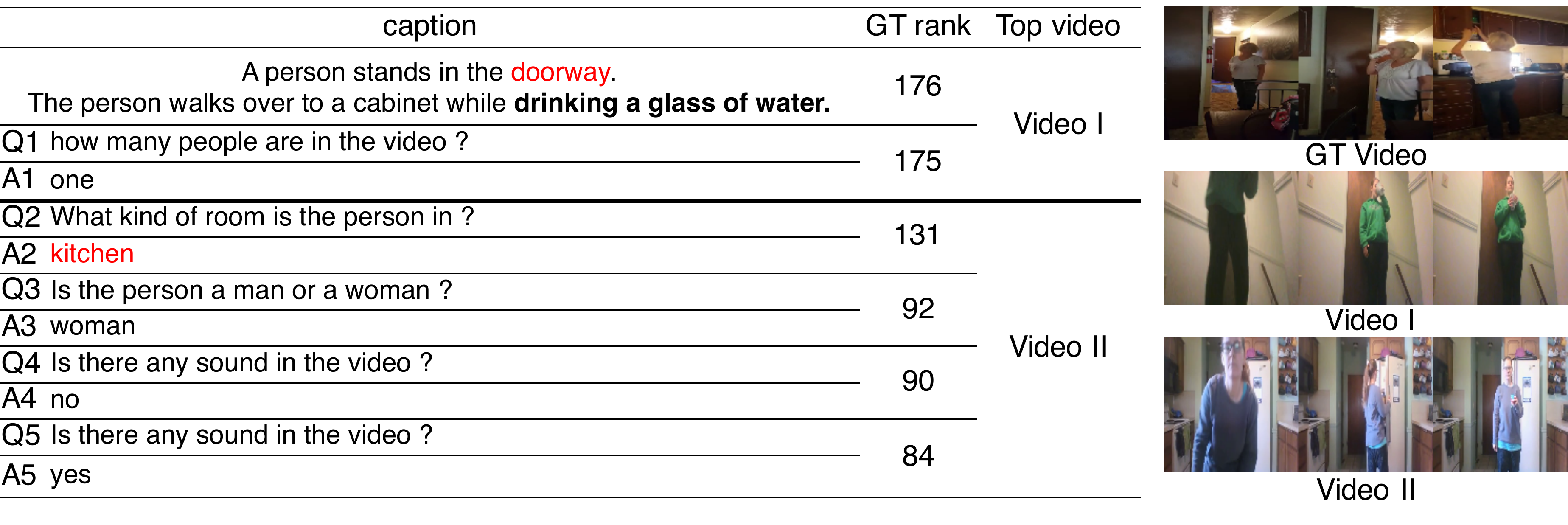}
 %\caption{Qualitative results. The upper part shows the video targeted left is the dialog and the right is the top ranked video.}
 \caption{An example of qualitative results. The left part shows the dialog targeting the GT video and the right is the frames of the GT video and the top ranked videos. Both Video I and Video II in the figure reflect "drinking a glass of water" in the caption. The background of the top videos changes from the doorway to the kitchen as a result of the dialog.}
 \label{fig:qual_proposed_dialog}
\end{figure}

\noindent{\bf Qualitative Results.}
\figref{fig:qual_proposed_dialog} is the qualitative result of the proposed method when user study is performed. It is apparent that the contents of the dialog influence the retrieval result.

\section{Conclusion}
The purpose of this study was to introduce an interaction (dialog) in which the system generates a question based on the videos to be retrieved and the dialog history with the user, enabling the retrieval of the target video among similar ones. To achieve this, we proposed a model that can ask questions and utilize the dialog history. We showed its effectiveness in experiments using the AVSD dataset, with videos that are rich in scene developments.
In future, we could potentially improve the retrieval performance by constructing a network to handle the video features more effectively than pooling and concatenation.

\section{Acknowledgement}
This work was supported by JST CREST Grant Number JPMJCR1403, Japan. 

\bibliography{egbib}

\begin{thebibliography}{25}
\providecommand{\natexlab}[1]{#1}
\providecommand{\url}[1]{\texttt{#1}}
\expandafter\ifx\csname urlstyle\endcsname\relax
  \providecommand{\doi}[1]{doi: #1}\else
  \providecommand{\doi}{doi: \begingroup \urlstyle{rm}\Url}\fi

\bibitem[Alamri et~al.(2019)Alamri, Cartillier, Das, Wang, Lee, Anderson, Essa,
  Parikh, Batra, Cherian, Marks, and Hori]{alamri2019audiovisual}
Huda Alamri, Vincent Cartillier, Abhishek Das, Jue Wang, Stefan Lee, Peter
  Anderson, Irfan Essa, Devi Parikh, Dhruv Batra, Anoop Cherian, Tim~K. Marks,
  and Chiori Hori.
\newblock {Audio-Visual Scene-Aware Dialog}.
\newblock In \emph{CVPR}, 2019.

\bibitem[Carreira and Zisserman(2017)]{carreira2017quo}
Joao Carreira and Andrew Zisserman.
\newblock {Quo Vadis, Action Recognition? A New Model and the Kinetics
  Dataset}.
\newblock In \emph{CVPR}, 2017.

\bibitem[Das et~al.(2017{\natexlab{a}})Das, Kottur, Gupta, Singh, Yadav, Moura,
  Parikh, and Batra]{visdial}
Abhishek Das, Satwik Kottur, Khushi Gupta, Avi Singh, Deshraj Yadav,
  Jos\'e~M.F. Moura, Devi Parikh, and Dhruv Batra.
\newblock {V}isual {D}ialog.
\newblock In \emph{CVPR}, 2017{\natexlab{a}}.

\bibitem[Das et~al.(2017{\natexlab{b}})Das, Kottur, Moura, Lee, and
  Batra]{das2017visdialrl}
Abhishek Das, Satwik Kottur, Jos\'e~M.F. Moura, Stefan Lee, and Dhruv Batra.
\newblock {Learning Cooperative Visual Dialog Agents with Deep Reinforcement
  Learning}.
\newblock In \emph{ICCV}, 2017{\natexlab{b}}.

\bibitem[Deng et~al.(2009)Deng, Dong, Socher, Li, Li, and
  Fei-Fei]{deng2009imagenet}
Jia Deng, Wei Dong, Richard Socher, Li-Jia Li, Kai Li, and Li~Fei-Fei.
\newblock {ImageNet: A Large-Scale Hierarchical Image Database}.
\newblock In \emph{CVPR}, 2009.

\bibitem[Dong et~al.(2016)Dong, Li, and Snoek]{dong2016word2visualvec}
Jianfeng Dong, Xirong Li, and Cees~GM Snoek.
\newblock {Word2visualvec: Image and Video to Sentence Matching by Visual
  Feature Prediction}.
\newblock \emph{arXiv preprint arXiv:1604.06838}, 2016.

\bibitem[Gemmeke et~al.(2017)Gemmeke, Ellis, Freedman, Jansen, Lawrence, Moore,
  Plakal, and Ritter]{gemmeke2017audio}
Jort~F Gemmeke, Daniel~PW Ellis, Dylan Freedman, Aren Jansen, Wade Lawrence,
  R~Channing Moore, Manoj Plakal, and Marvin Ritter.
\newblock {AUDIO SET: AN ONTOLOGY AND HUMAN-LABELED DATASET FOR AUDIO EVENTS}.
\newblock In \emph{ICASSP}, 2017.

\bibitem[He et~al.(2016)He, Zhang, Ren, and Sun]{he2016deep}
Kaiming He, Xiangyu Zhang, Shaoqing Ren, and Jian Sun.
\newblock {Deep Residual Learning for Image Recognition}.
\newblock In \emph{CVPR}, 2016.

\bibitem[Hendricks et~al.(2017)Hendricks, Wang, Shechtman, Sivic, Darrell, and
  Russell]{Hendricks2017LocalizingMI}
Lisa~Anne Hendricks, Oliver Wang, Eli Shechtman, Josef Sivic, Trevor Darrell,
  and Bryan~C. Russell.
\newblock {Localizing Moments in Video with Natural Language}.
\newblock In \emph{ICCV}, 2017.

\bibitem[Hershey et~al.(2017)Hershey, Chaudhuri, Ellis, Gemmeke, Jansen, Moore,
  Plakal, Platt, Saurous, Seybold, Slaney, Weiss, and Wilson]{Hershey2017CNNAF}
Shawn Hershey, Sourish Chaudhuri, Daniel P.~W. Ellis, Jort~F. Gemmeke, Aren
  Jansen, R.~Channing Moore, Manoj Plakal, Devin Platt, Rif~A. Saurous, Bryan
  Seybold, Malcolm Slaney, Ron~J. Weiss, and Kevin~W. Wilson.
\newblock {CNN ARCHITECTURES FOR LARGE-SCALE AUDIO CLASSIFICATION}.
\newblock In \emph{ICASSP}, 2017.

\bibitem[Hori et~al.(2018)Hori, Alamri, Wang, Winchern, Hori, Cherian, Marks,
  Cartillier, Gontijo~Lopes, Das, Essa, Batra, and Parikh]{hori2018end}
Chiori Hori, Huda Alamri, Jue Wang, Gordon Winchern, Takaaki Hori, Anoop
  Cherian, Tim Marks, Vincent Cartillier, Raphael Gontijo~Lopes, Abhishek Das,
  Irfan Essa, Dhruv Batra, and Devi Parikh.
\newblock {End-to-End Audio Visual Scene-Aware Dialog using Multimodal
  Attention-Based Video Features}.
\newblock \emph{arXiv preprint arXiv:1806.08409}, 2018.

\bibitem[Jang et~al.(2017)Jang, Song, Yu, Kim, and Kim]{jang-CVPR-2017}
Yunseok Jang, Yale Song, Youngjae Yu, Youngjin Kim, and Gunhee Kim.
\newblock {TGIF-QA: Toward Spatio-Temporal Reasoning in Visual Question
  Answering}.
\newblock In \emph{CVPR}, 2017.

\bibitem[Kay et~al.(2017)Kay, Carreira, Simonyan, Zhang, Hillier,
  Vijayanarasimhan, Viola, Green, Back, Natsev, et~al.]{kay2017kinetics}
Will Kay, Joao Carreira, Karen Simonyan, Brian Zhang, Chloe Hillier, Sudheendra
  Vijayanarasimhan, Fabio Viola, Tim Green, Trevor Back, Paul Natsev, et~al.
\newblock {The Kinetics Human Action Video Dataset}.
\newblock \emph{arXiv preprint arXiv:1705.06950}, 2017.

\bibitem[Kingma and Ba(2015)]{kingma:adam}
Diederick~P Kingma and Jimmy Ba.
\newblock {Adam: A Method for Stochastic Optimization}.
\newblock In \emph{ICLR}, 2015.

\bibitem[Miech et~al.(2018)Miech, Laptev, and Sivic]{miech18learning}
Antoine Miech, Ivan Laptev, and Josef Sivic.
\newblock Learning a {T}ext-{V}ideo {E}mbedding from {I}ncomplete and
  {H}eterogeneous {D}ata.
\newblock \emph{arXiv preprint arXiv:1804.02516}, 2018.

\bibitem[Mithun et~al.(2018)Mithun, Li, Metze, and
  Roy-Chowdhury]{mithun2018learning}
Niluthpol~Chowdhury Mithun, Juncheng Li, Florian Metze, and Amit~K
  Roy-Chowdhury.
\newblock {Learning Joint Embedding with Multimodal Cues for Cross-Modal
  Video-Text Retrieval}.
\newblock In \emph{ICMR}, 2018.

\bibitem[Otani et~al.(2016)Otani, Nakashima, Rahtu, Heikkil{\"a}, and
  Yokoya]{otani2016learning}
Mayu Otani, Yuta Nakashima, Esa Rahtu, Janne Heikkil{\"a}, and Naokazu Yokoya.
\newblock {Learning Joint Representations of Videos and Sentences with Web
  Image Search}.
\newblock In \emph{ECCV}, 2016.

\bibitem[Pasunuru and Bansal(2018)]{pasunuru2018game}
Ramakanth Pasunuru and Mohit Bansal.
\newblock {Game-Based Video-Context Dialogue}.
\newblock In \emph{EMNLP}, 2018.

\bibitem[Shao et~al.(2018)Shao, Xiong, Zhao, Huang, Qiao, and
  Lin]{shao2018find}
Dian Shao, Yu~Xiong, Yue Zhao, Qingqiu Huang, Yu~Qiao, and Dahua Lin.
\newblock {Find and Focus: Retrieve and Localize Video Events with Natural
  Language Queries}.
\newblock In \emph{ECCV}, 2018.

\bibitem[Sigurdsson et~al.(2016)Sigurdsson, Varol, Wang, Farhadi, Laptev, and
  Gupta]{sigurdsson2016hollywood}
Gunnar~A Sigurdsson, G{\"u}l Varol, Xiaolong Wang, Ali Farhadi, Ivan Laptev,
  and Abhinav Gupta.
\newblock {Hollywood in Homes: Crowdsourcing Data Collection for Activity
  Understanding}.
\newblock In \emph{ECCV}, 2016.

\bibitem[Wang et~al.(2016)Wang, Yin, Wang, Wu, and Wang]{wang2016comprehensive}
Kaiye Wang, Qiyue Yin, Wei Wang, Shu Wu, and Liang Wang.
\newblock {A Comprehensive Survey on Cross-modal Retrieval}.
\newblock \emph{arXiv preprint arXiv:1607.06215}, 2016.

\bibitem[Yamaguchi et~al.(2017)Yamaguchi, Saito, Ushiku, and
  Harada]{yamaguchi2017spatio}
Masataka Yamaguchi, Kuniaki Saito, Yoshitaka Ushiku, and Tatsuya Harada.
\newblock {Spatio-temporal Person Retrieval via Natural Language Queries}.
\newblock In \emph{ICCV}, 2017.

\bibitem[Yang et~al.(2018)Yang, Zhang, and Xu]{yang2018text2video}
Xiaoshan Yang, Tianzhu Zhang, and Changsheng Xu.
\newblock {Text2Video: An End-to-end Learning Framework for Expressing Text
  with Videos}.
\newblock \emph{IEEE Transactions on Multimedia}, 20\penalty0 (9):\penalty0
  2360--2370, 2018.

\bibitem[Yu et~al.(2018)Yu, Kim, and Kim]{yu2018joint}
Youngjae Yu, Jongseok Kim, and Gunhee Kim.
\newblock {A Joint Sequence Fusion Model for Video Question Answering and
  Retrieval}.
\newblock In \emph{ECCV}, 2018.

\bibitem[Zhao et~al.(2018)Zhao, Jiang, Cai, Xiao, He, and Pu]{zhao2018multi}
Zhou Zhao, Xinghua Jiang, Deng Cai, Jun Xiao, Xiaofei He, and Shiliang Pu.
\newblock {Multi-Turn Video Question Answering via Multi-Stream Hierarchical
  Attention Context Network.}
\newblock In \emph{IJCAI}, 2018.

\end{thebibliography}
\end{document}